\documentclass[journal]{IEEEtran}

\ifCLASSINFOpdf
   \usepackage[pdftex]{graphicx}
  
\else
  
 \usepackage{graphicx}
  
\fi

\usepackage{amsmath}
\usepackage{amsfonts} 
\usepackage{balance}
 \usepackage{xcolor}
\usepackage{caption,subcaption}
\usepackage{arydshln}
\usepackage{amsthm}
\newtheorem{definition}{Definition}
\usepackage[utf8]{inputenc}
\DeclareUnicodeCharacter{2212}{-}
\begin{document}

\title{\textit{TextConvoNet}: A Convolutional Neural Network based Architecture for Text Classification}

\author{Sanskar Soni, Satyendra Singh Chouhan, Santosh Singh Rathore

\thanks{Sanskar Soni and Satyendra Singh Chouhan is with the Department
of Computer Science and Engineeering, Malaviya National Institute of Technology, Jaipur, 302017, INDIA E-mail: \{2018ucp1265,sschouhan.cse\}@mnit.ac.in}
 \thanks{Santosh Singh Rathore is with the Department of Information Technology, ABV-IIITM Gwalior,474015, INDIA E-mail: santoshs@iiitm.ac.in} 
}

\maketitle

\begin{abstract}
In recent years, deep learning-based models have significantly improved the Natural Language Processing (NLP) tasks. Specifically, the Convolutional Neural Network (CNN), initially used for computer vision, has shown remarkable performance for text data in various NLP problems. Most of the existing CNN-based models use 1-dimensional convolving filters (\textit{n-gram} detectors), where each filter specialises in extracting \textit{n-grams} features of a particular input word embedding. The input word embeddings, also called sentence matrix, is treated as a matrix where each row is a word vector. Thus, it allows the model to apply one-dimensional convolution and only extract \textit{n-gram} based features from a sentence matrix. These features can be termed as \textit{intra-sentence n-gram features}. To the extent of our knowledge, all the existing CNN models are based on the aforementioned concept. In this paper, we present a CNN-based architecture \textit{TextConvoNet} that not only extracts the intra-sentence n-gram features but also captures the inter-sentence n-gram features in input text data. It uses an alternative approach for input matrix representation and applies a two-dimensional multi-scale convolutional operation on the input. To evaluate the performance of \textit{TextConvoNet}, we perform an experimental study on five text classification datasets. The results are evaluated by using various performance metrics. The experimental results show that the presented \textit{TextConvoNet} outperforms state-of-the-art machine learning and deep learning models for text classification purposes.

\end{abstract}

\begin{IEEEkeywords}
Text classification, CNN, Multi-dimensional Convolution, Deep learning.
\end{IEEEkeywords}

\IEEEpeerreviewmaketitle

\section{Introduction}

Natural language processing (NLP) involves computational processing and understanding of the natural/human languages. It involves various tasks that rely on various statistics and data-driven computation techniques~\cite{jones1994natural}. One of the important tasks in NLP is text classification. It is a classical problem where the prime objective is to classify (assign labels or tags) to the textual contents~\cite{kowsari2019text}. Textual contents can either be sentences, paragraphs, or queries~\cite{kowsari2019text,minaee2020deep}. There are many real-world applications of text classification such as sentiment analysis~\cite{wang2012baselines}, news classification~\cite{bozarth2020toward}, intent classification~\cite{parmar2020owi}, spam detection~\cite{wang2019drifted}, and so on.

Text classification can be done by manual labeling of the textual data. However, with the exponential growth of text data in the industry and over the Internet, automated text categorization has become very important. Automated text classification approaches can be broadly classified into three categories: Rule-based, Data-Driven based (Machine Learning/Deep Learning-based approaches), and Hybrid approaches. Rule-based approaches classify text into different categories using a  set of pre-defined rules. However, it requires complete domain knowledge~\cite{scott1999feature,hadi2018integrating}. Alternatively, machine learning-based approaches have proven to be significantly effective in recent years. All the machine learning approaches work in two stages: first, they extract some handcrafted features from the text. Next, these features are fed into a machine learning model. For extracting the handcrafted features, a bag of words, \textit{n-grams} based model, term frequency, and inverse document frequency (TF-IDF) and their extensions were popularly used. For the second stage, many classical Machine Learning algorithms such as Support Vector Machine (SVM), Decision Tree (DT), Conditional Probability-based such as Naïve Bayes, and other Ensemble-based approaches are used~\cite{joachims1998text,ikonomakis2005text,hacohen2020influence,mccallum1998comparison}.

Recently, some of the Deep Learning methods, specifically RNN (Recurrent Neural Network) and CNN (Convolutional Neural Network), have shown remarkable results in text classification~\cite{wang2017combining,zhou2015c,adhikari2019rethinking,yang2018sgm,duque2019squeezed,liu2020multichannel,minaee2021deep}. CNN-based models are trained to recognize patterns in text, such as key phrases. Most CNN-based models utilize one-dimensional (1-D) convolution followed by a one-dimensional max-pooling operation to extract a feature vector from the input word embeddings. This feature vector is fed into the classification layer as an input for classification purposes. Input word embedding is a word matrix where each row represents a word vector. Therefore, one-dimensional (1-D) convolution extracts \textit{n-gram} based features by performing convolution operation on two or more than two-word vectors at a time.   

However, improving text classification results by utilizing the \textit{n-gram} features in between different sentences using convolution operation still remains an open research question for all the researchers. Furthermore, the input matrix structure also remains a point to ponder, which could be revamped to apply multidimensional convolution. This paper presents, \textit{TextConvoNet},  a new CNN-based architecture for text classification. Unlike existing works, the proposed architecture uses a 2-dimensional convolutional filter to extract the intra-sentence and inter-sentence \textit{n-gram} features from text data. First, it represents the text data as a paragraph level (multi-sentence) embedding matrix, which helps in applying 2-dimensional convolutional filters. Thereafter, multiple convolutional filters are applied to the extract features. The resultant features are concatenated and fed into the classification layer for classification purposes. 
To evaluate the performance of the presented \textit{TextConvoNet}, we perform experiments on some benchmark binary as well as multi-class classification datasets. Evaluation of the \textit{TextConvoNet} is done based on various performance metrics such as accuracy, precision, recall, F1-score, specificity, and G-mean. Additionally, we compare the performance of the \textit{TextConvoNet} with state-or-the-art classification models. 

\subsection{Contributions}
The contributions of this paper are as follows.

\begin{itemize}
    \item An approach is presented to represent input text data as a  multidimensional word embedding representation. It would overcome the restriction of applying one-dimensional convolution only.
    \item A multi-scale feature extraction approach using two-dimensional convolution and pooling operation is presented.
    \item The overall model is tested on five benchmark datasets, and extensive experiments are performed to test the proposed model’s effectiveness.
    \item We have also evaluated different possible versions of \textit{TextConvoNet} for text classification purpose.
    \item An extensive comparison of the proposed model with existing state-of-the-art CNN-based models on five benchmark datasets is performed to show the model’s efficacy.
    \item Ablation study of \textit{TextConvoNet} is performed, which is focused on optimizing the hyper-parameters involved at different layers used in the presented model.
\end{itemize}

The rest of the article is organized as follows. Section II discusses the literature review. Section III starts with background information on text classification using CNN and then provides details of the presented \textit{TextConvoNet} architecture. Section IV provides the details of the experimental setup and analysis. It starts with details of datasets used followed by performance metrics, implementation details, experimental results, and comparison with the state-of-the-art. It also presents the ablation study and few-shot learning analysis of \textit{TextConvoNet}. Section V presents the conclusions along with the future directions.

\section{Literature Review}
Text classification is one of the important tasks in Natural Language Processing. There have been many data-driven-based approaches suggested for text classification. Recently, deep learning-based approaches have emerged and performed significantly well in text classification.  In this section, we discuss some of the relevant deep learning-based models suggested for text classification.

There are mainly two neural networks that have been very popular in NLP problems: Long Short Term Memory (LSTM) and Convolutional Neural Network (CNN). LSTM can extract current information and remember the past data points in sequence~\cite{adhikari2019rethinking,yang2018sgm,zhou2016attention,yang2016hierarchical}. However, LSTM-based models have very high training time because it is characterized as each token per step. In addition, LSTM with attention networks introduces an additional computation burden. It is because of the exponential function and normalized alignment score computation of all the words in the text~\cite{wang2016attention}. 
 
One of the early CNN-based models was Dynamic CNN (DCNN) for text classification~\cite{kalchbrenner2014convolutional}. It used dynamic max-pooling, where the first layer of DCNN makes a sentence matrix using the word embeddings. Then it has a convolutional architecture that uses repeated convolutional layers with dynamic k-max-pooling to extract feature maps on the sentence. These features maps are capable enough of capturing the short and long-range relationship between the words.

Later on, Kim~\cite{kim2014} gave the simplest CNN-based model for text classification, which has become the benchmark architecture for many recent models for text classification. Kim’s model adapted a single layer of convolution on top of the input word vectors obtained from an unsupervised neural language model (word2vec). It has been evident to improve upon the state-of-the-art binary and multi-class text classification problems. Recently, there have been some attempts towards improving the architectures of CNN-based models~\cite{liu2017deep,johnson2014effective,johnson2017deep,duque2019squeezed,liu2020multichannel}. Instead of using pre-trained low-dimensional word vectors as an input to CNN, authors in~\cite{johnson2017deep} directly applied CNNs to high-dimensional text data to learn the embeddings of small text regions for the classification. Most of the existing works have shown that convolutions of sizes 2, 3, 4, or 5 give significant results in text classification.

Some of the researches explored the impact in model performance while using the word embeddings and CNN architectures. Encouraged by VGG~\cite{simonyan2014very} and ResNets~\cite{he2016deep}, Conneau et al.~\cite{conneau2017very} proposed a VDCNN (Very Deep CNN) model for text processing. It applies CNN directly at the character level and uses small convolutions and pooling functions. The research exhibited that the performance of VDCNN improves with the increase in depth. 
In another work, 

Le et al.~\cite{le2018convolutional}  shown that deep architectures can outperform shallow architectures where text data is represented as a sequence of characters. Later on, in another research, Squeezed-VDCNN suggested ~\cite{duque2019squeezed}, which improved VDCNN to work on mobile platforms.  However, a basic shallow-and-wide network outperforms deep models, such as DenseNet~\cite{huang2017densely}, with word embeddings as inputs.

\begin{figure*}[htb]
  \centering
    \includegraphics[width=6.2in,height=3.7in,angle=0]{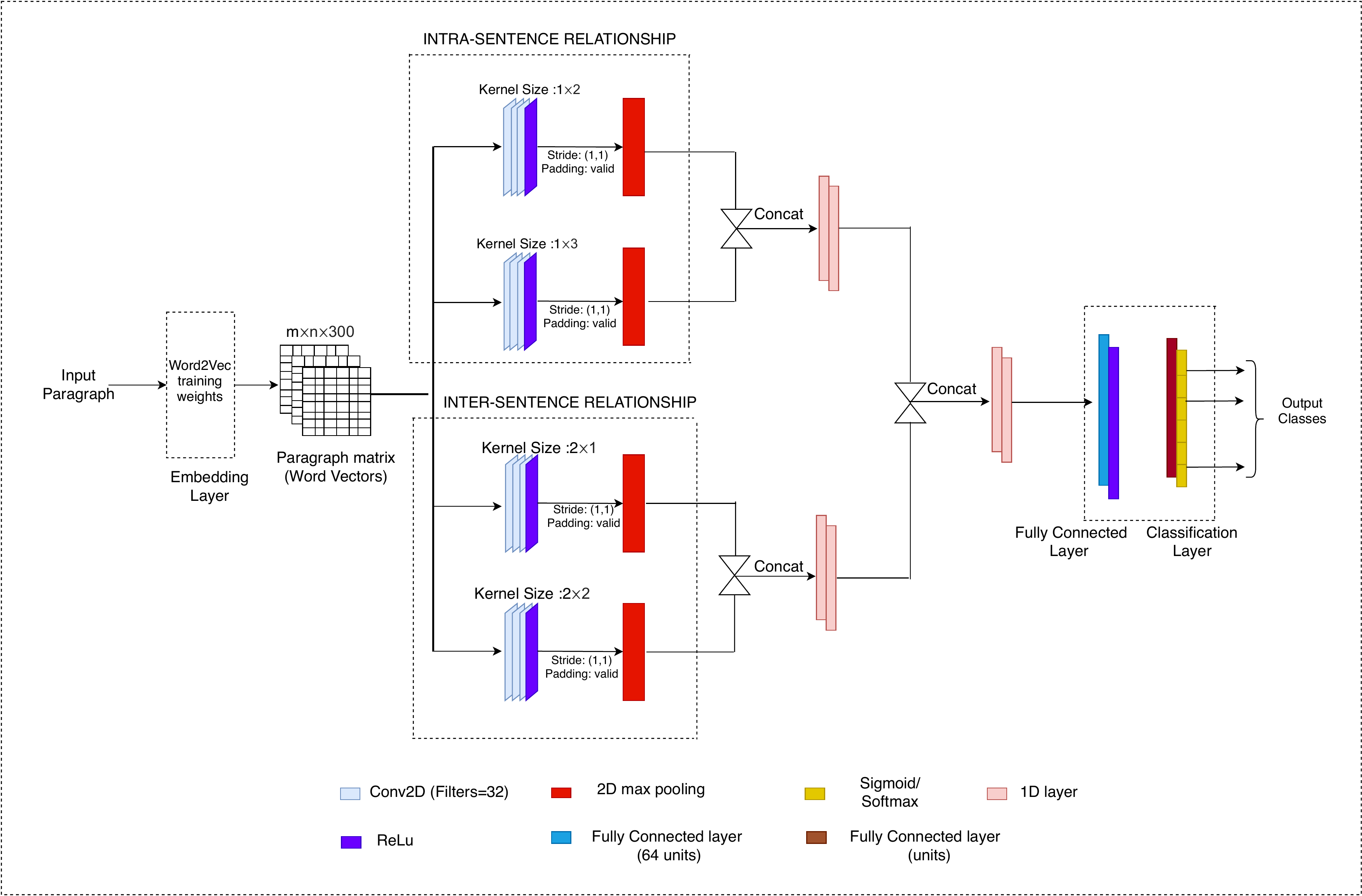}
  \caption{Proposed \textit{TextConvoNet} Architecture}
  \label{fig1}
\end{figure*}
\begin{figure}[htb]
\centering
\includegraphics[width=3.5in,height=1.5in,angle=0]{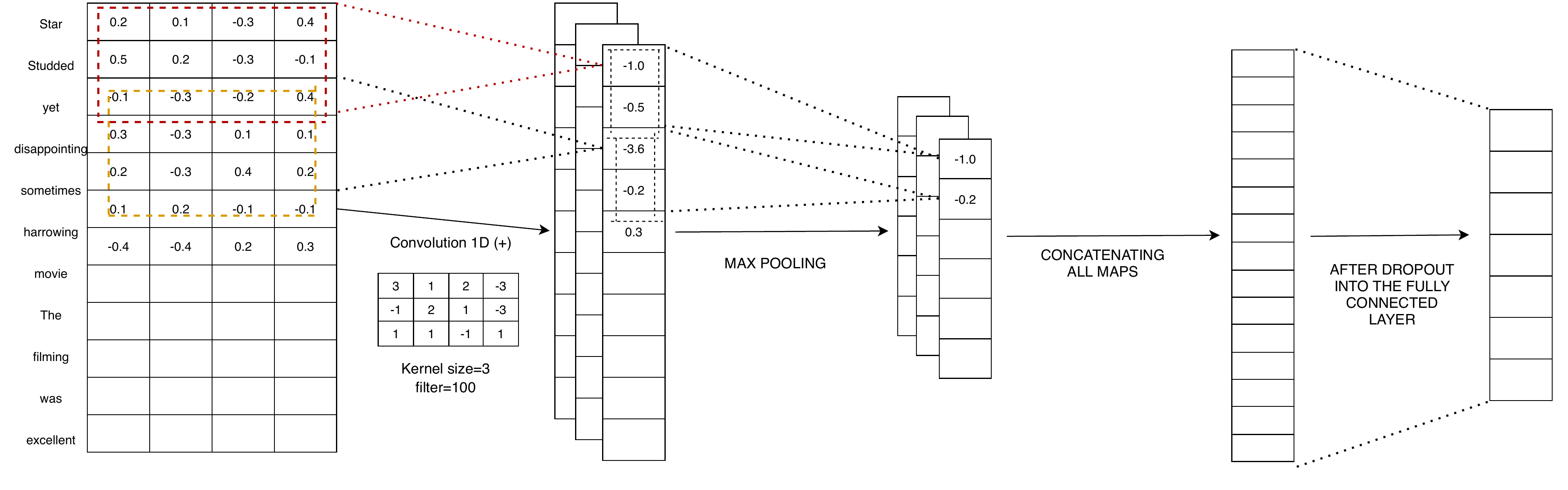}
\caption{Example: Text classification using CNN\cite{kim2014}}
\label{example1}
\vspace{-3.0ex}
\end{figure}

To the best of our knowledge, above discussed, all the  CNN-based networks extract the \textit{n-gram} based feature using varied sizes of kernels/filters. In light of the above works, we present a novel CNN-based architecture that extracts intra-sentence n-gram features and captures the inter-sentence n-gram features.

\section{Proposed TextConvoNet}
In this section, first, we discuss the existing CNN-based approach of text classification taking a simple example. Next, we present the proposed \textit{TextConvoNet} framework.  

\subsection{Text classification using CNN}\label{3(A)}
Text classification problems can be formally defined as follows.
\begin{definition}
Given a text dataset $\mathcal{T}$ consisting of labelled text articles. Depending on a particular NLP task, text articles have a particular label/class $l \in L$. In case of binary classification, there will be total of two labels for the text dataset. A text article $te \in \mathcal{T}$ consists of sentences and words. Let us say, text article $te_{i}$ contain $m$ sentences ${s_{1},\ldots,s_{m}}$ and sentence $s_{j} \ (0 \leq j \leq m)$ contain $n$ words. Words in a sentence $s_{j}$ can be represented as $W_{j}$ = ${w_{1}, \ldots, w_{n}}$. 
 \end{definition}
The objective of text classification is to learn a model $M$ that can correctly classify any new text articles $t_{new}$ into label $l \in L$. 

Kim et al.,~\cite{kim2014} presented a simple and effective architecture for the text classification. We call it Kim's CNN model throughout the paper for simplicity. This presented architecture served as a guiding light and basis for many CNN-based architectures for text classification. Many recent architectures internally use this model~\cite{jacovi2018understanding,wang2019clinical,kim2019sentiment,akhter2020document}. In Kim's CNN model, sentences are matched to the embedding vectors made available as an input matrix to the model. It makes use of only a single layer of convolution with word vectors on top obtained from an existing pre-trained model, with kernel sizes 3, 4, and 5. The resultant feature maps are further processed using a max-pooling layer to distill or summarize the extracted features, subsequently sent to a fully connected layer. Figure~\ref{example1} shows a simple example of text classification using CNN-based Kim's model.

As shown in Figure~\ref{example1}, the input to the model is a sentence represented as a matrix. Each row of the matrix is a vector that represents a word. 1D convolution is performed onto the matrix with the kernel size being 3 along with 4 and 5. Max-Pooling is performed upon the filter maps, which are further concatenated and sent to the last fully connected layer for the classification purpose. Formally, the sentence modeling is as follows.
\vspace{2.0pt}

\noindent\textbf{Sentence Modelling}. For each sentence, $w_p \in \mathbb{R}_z$ denote the word embedding for the $p^{th}$ word in the sentence, where $z$ is the word embedding dimension. Suppose that a sentence has $n$ words, the sentence can now be represented as an embedding matrix $W \in \mathbb{R}^{n\times z}$. So we can refer to it as a word matrix where every row denotes the vectors for a particular word of the sentence. Let $w_{p:p+q}$ represents the concatenation of vectors $w_{p},w_{p+1},\ldots,w_q$. The convolution operation is performed on this input embedding layer. It involves a filter $k\in \mathbb{R}^{s,z}$ that applies to a window of $s$ words to produce a new feature. For example, a feature ${c_{p}}$ is generated using the window of words $w_{p:p+s−1}$ by Equation \ref{eq1}.

\begin{equation}\label{eq1}
    c_{p} = f(w_{p:p+s-1}.k + b) 
\end{equation}

Here, $b\in R$ and \textit{f} denotes the bias and non-linear activation function respectively. The filter (kernel) \textit{k} applies to all possible windows using the same weights to create the feature map.
\begin{equation}
    	c = [c_{1}, c_{2}, \ldots, c_{n-s+1}]
\end{equation}
 	
\subsection{Proposed \textit{TextConvoNet} Architecture}
In real-world scenarios, the paragraphs are stringed together in a very complex manner, making it very difficult for any model to come up with correct labeling, whether it be any sentiment or a news category. So there may be instances when the model cannot extract the inter-sentence features and hence fails to come up with a suitable result. Taking inspiration from the above shortcoming, we try to present an alternative input structure for the model and propose a novel CNN model using the alternative input structure and employing 2-D Convolution \cite{merdivan2019image}. The description of our input structure is mentioned below.
\vspace{0.5pt}

\noindent\textbf{Paragraph Modelling}. For each sentence in a paragraph, let $w_{i} \in \mathbb{R}^z$ represents the word embedding for the $i^{th}$ word in the sentence, where \textit{z} is the dimension of the word embedding. Given that a paragraph has \textit{m} sentences and \textit{n} word in each sentence, the paragraph can be represented as an embedding matrix $W \in \mathbb{R}^{m\times n \times d}$.
Such type of arrangement could be termed as a paragraph matrix where each row depicts each sentence of a paragraph, with each cell as a single word \textit{w} and the $3^{rd}$ dimension as the embeddings or the word vectors. 

The overall architecture of our proposed model, \textit{TextConvoNet}, is shown in Figure~\ref{fig1}. The presented \textit{TextConvoNet} model uses an alternate input structure of the paragraph, using 2D convolution instead of 1D convolution and differing kernel sizes. \textit{TextConvoNet} sends the input matrix into 4 parallel pathways of Convolution layers. The first two layers (intra-sentence layers) with 32 filters each and kernel sizes of ($1\times2$ and $1 \times 3$), respectively, are concatenated and have the role of extricating the intra-sentence n-gram features. The other two layers (inter-sentence) with 32 filters each and kernel sizes of ($2\times1$ and $2\times2$) concatenated together have the sole purpose of drawing out the inter-sentence n-gram features. These two intra-sentence and inter-sentence layers are further concatenated and fed into the fully connected layer consisting of 64 neurons and subsequently perform the relevant classification task. A detailed explanation of the architecture is given as follows. 

\subsubsection{Convolution Layer}
This layer applies filters to input to create feature maps and condense out the input's detected features. Let $F(m,n)$ be an input paragraph of size $m \times n$ and \textit{f} is the filter with kernel size of $(g \times h)$ where $b(g,h)$ represents the biases. The outcome of the convolutional layer is computed by Equation \ref{eq3}.

\begin{equation}\label{eq3}
    R(m,n)=f_{w}^{(g\times h )}\otimes F(m,n)+b^{(g,h)}
\end{equation}

\subsubsection{ ReLu Activation Layer}
The purpose of the ReLu activation layer after each convolution layer is to normalize output. This layer also aids the model to learn something complex and complicated with a reduced possibility of vanishing gradient and cheap computation costs. The activation function for ReLu is calculated using Equation \ref{eq4}. Here $\gamma$ is the input to the ReLu function. 

\begin{equation}\label{eq4}
    f(\gamma)=max(0,\gamma )
\end{equation}

\subsubsection{ Concatenation Layer}
It concatenates the various input blobs and concatenates them in a continuous manner.

\subsubsection{Fully Connected Layer} This is a multilayer perceptron that is connected to all the activations from the previous layers. The activation of these neurons is calculated by matrix multiplication of its weights added by an offset value.

Let assume the input to be ${\beta}$ with size \textit{u} and \textit{r} be the number of neurons in the fully connected layer. The output of this layer is computed using Equation \ref{Eq5}.

\begin{equation}\label{Eq5}
    c(\beta)=\vartheta (U\ast \beta)
\end{equation}
where $\vartheta$ denotes the activation function and output matrix is $F_{u \times r}$.

\subsubsection{Dropout Layer} The dropout layer randomly activates or deactivates (make them 0) the outgoing edges of hidden units at each update of the training phase, which also helps to reduce overfitting.

\subsubsection{Classification Layer} This is the final layer that performs classification based on the attributes extricated by the previous layers. It is a traditional ANN layer with softmax or sigmoid as the activation function.

\subsubsection{Loss Function}
For binary-class text classification task, \textit{TextConvoNet} is trained by minimizing the binary-cross entropy (Equation \ref{Eq6}) over a sigmoid activation function. For the task of multi-class classification, \textit{TextConvoNet} is trained by minimizing the categorical-cross entropy (Equation \ref{Eq7}) over a softmax activation function. The above loss functions can be formulated as
\begin{equation}\label{Eq6}
\scriptsize
BCE =-\frac{1}{m}\sum_{i}^{m} \sum_{j}^{c}y_{ij} log(\sigma(t_{ij}))) - (1 - y_{ij}) log(1 - \sigma(t_{ij})))
\end{equation}

\begin{equation}\label{Eq7}
CCE = -\frac{1}{m}\sum_{i}^{m} \sum_{j}^{c}y_{ij}log\left ( \frac{e^{t_{ij}}}{\sum_{r=1}^{c} e^{t_{ij}}}\right )
\end{equation}

Here \textit{i} is the index of a training instance, \textit{j} is the index of a label (class), $t_{ij}$ is  output of the final fully connected layer, and $y_{ij}$ is the ground truth (actual value) of $i^{th}$ training sample
of the $j^{th}$ class.

\subsection{Analysis of \textit{TextConvoNet}}
Since almost all the real-life conversations, reviews, remarks are generally very long and complex and thus convey a different perspective in each line, although there is only a single deep-rooted sentiment attached to the whole paragraph. To uphold the semantics, the paragraph is converted into paragraph-level sentence embedding without any preprocessing of text.  The Embedding matrix is then sent into 4 lateral pathways subdivided into the the intra-sentence (kernel sizes $1\times 2$,$1\times3$) layer and the inter-sentence layer (kernel sizes $2\times1$, $2\times2$) with 32 filters in every layer. These hyperparameters were selected through the GridSearchCV method from the plethora of other suitable hyperparameter choices. The results also complemented our thinking/approach of selecting small window sizes to capture each and every minute detail. Similarly, using the GridSearchCV method, the learning rate was chosen to be $0.01$ and the number of neurons to be 64 for the final fully connected layer. The convolutional layers were limited to four only, as increased layers led to overfitting.


\subsection{Variants of TextConvoNet}
We have created various variants of the proposed model to come up with an effective text classification framework.  Figure \ref{fig1} shows the baseline model. However, it might be interesting to see whether increasing the number of \textit{n-gram} based kernels, and inter-sentence kernels will improve the efficacy of the model. Thus, we have extended the baseline to create two versions of \textit{TextConvoNet}: \textit{TextConvoNet\_4} and \textit{TextConvoNet\_6}. 

\begin{itemize}
\item \textit{TextConvoNet\_4}: Our base/parent model with 4 convolution layers (with different kernel sizes), 2 for extracting out the intra-sentence \textit{n-gram} features, and the other 2 for extricating the \textit{n-gram} based inter-sentence attributes. 

\item \textit{TextConvoNet\_6}: Same framework as above but extending the convolutional pathways to 6, 3 for drawing out intra-sentence \textit{n-gram} features and other 3 for inter-sentential \textit{n-gram} features. 
\end{itemize}

We have also performed modifications on various parameters of \textit{TextConvoNet}: number of filters, dropout rate, kernel sizes, number of nodes in fully connected layer and optimizers, etc. However, the effectiveness of these modifications is experimentally validated in section \ref{results}.   

\begin{table*}[htb]
\caption{Details of the used experimental datasets}
\scriptsize
\vspace{-1.0ex}
\label{datasetdetails}
\centering
\begin{tabular}{lllllll}
\hline
\textbf{Dataset} & \textbf{Description} & \textbf{\#Classes} & \textbf{Exceeding Ratio} & \textbf{Vocabulary size} & \textbf{\#Training set} & \textbf{\#Testing set} \\ \hline
DATASET-1        & BINARY             & 2                 & 0.471                    & 16449                    & 6911                    &  {1821}                   \\
 {DATASET-2}        &   {BINARY}              &  {2}                  &  {0.495}                    &  {39515}                    &  {15000}                    &  {5000}                   \\
 {DATASET-3}        &  {MULTICLASS}          &  {8}                  &  {0.297}                    &  {13737}                    &  {4937}                    &  {2175}                   \\
 {DATASET-4}        &  {MULTICLASS}          &  {3}                  &  {0.563}                    &  {16230}                    &  {10000}                   &  {3000}                   \\
 {DATASET-5}        &  {MULTICLASS}          &  {5}                  &  {0.553}                    &  {24742}                    &  {6000}                    &  {1500}                   \\ \hline
\end{tabular}
\\
 Note: Exceeding ratio$=$  number of samples with a length greater than the average length$/$number of all training samples.
 \vspace{-1.0ex}
\end{table*}

\section{Experimental Setup and Analysis}
In this section, first, we provide a description of the used datasets. For the performance evaluation, we conducted experiments on various binary and multi-class text classification datasets. This section also discusses the performance metrics used and baseline machine learning and deep learning models used for comparison. In the last, we discuss the results of the experimental analysis.

In experiment and results, first, we present the performance comparison of \textit{TextConvoNet} with other baseline models. Moreover, we conduct a statistical test to assess whether \textit{TextConvoNet} performed significantly differently from other baseline models. Next, we discuss the ablation study of \textit{TextConvoNet} with respect to various parameters of the presented model. In addition, we analyze the performance of our model by varying the number of sentences in a paragraph. Thereafter, we discuss the experimental results with respect to dataset size, i.e., we want to see how does the presented \textit{TextConvoNet} performs with minimal data and in challenging scenarios.   

\subsection{Used Datasets}\label{IV-A}
We have performed experiments on various publicly available binary and multiclass datasets. Only a subset of instances from the datasets was included for training and testing. The complete details about the datasets are given in Table~\ref{datasetdetails}. No additional changes have been done in the datasets, and no preprocessing has been applied to the text. For the experiments, we used 2 binary and 3 multiclass datasets. The binary datasets are the famous SST-2 and Amazon Review Dataset, whereas multiclass datasets consisted of Ohsumed(R8), Twitter Airline Sentiment, and the Coronavirus Tagged Datasets. All the datasets are publicly available and are sourced from Kaggle. The details of the datasets are mentioned below.

\begin{itemize}
\item DATASET-1 \footnote{https://www.kaggle.com/jgggjkmf/binary-sst2-dataset}:\textbf{Binary SST$-$2}: This dataset is similar to the Stanford Sentiment Treebank dataset with only positive and negative reviews, We have removed all neutral reviews from the dataset.

\item DATASET-2~\footnote{https://www.kaggle.com/bittlingmayer/amazonreviews}:\textbf{Amazon Review for Sentiment Analysis Dataset}: This dataset contains a few million customer reviews and the star ratings.

\item DATASET-3~\footnote{https://www.kaggle.com/weipengfei/ohr8r52}:\textbf{R8 Dataset}: This is a subset of Reuters21578 dataset containing 8 categories for multiclass classification.

\item DATASET-4~\footnote{https://www.kaggle.com/crowdflower/twitter-airline-sentiment}:\textbf{Twitter  User Airline Sentiment}: The dataset contains the tweets of six different airlines as positive, negative, or neutral.

\item DATASET-5~\footnote{https://www.kaggle.com/datatattle/covid-19-nlp-text-classification}: \textbf {Covid Tweets}: The tweets have been pulled from Twitter followed by manual tagging as Extremely Negative, Negative, Neutral, Positive and Extremely Positive.
\end{itemize}

 \begin{table*}[htb]
\caption{Description of the performance evaluation metrics}
\vspace{-1.0ex}
\label{PM}
\scriptsize
\renewcommand*{\arraystretch}{1.2}
\centering
\begin{tabular}{p{1.2cm}lp{11.0cm}}
\hline
\textbf{Measures}  & \textbf{Formula} & \textbf{Description} \\ \hline

Accuracy & $\frac{TP+TN}{TP+FP+TN+FN}$ & Accuracy refers to the amount of accurate assumptions the algorithm produces for forecasts of all sorts.\\   
Precision  & $ \frac{TP}{TP+FP}$ & Precision is the percentage of successful cases that were reported  correctly.  \\   

Recall &  $\frac{TP}{TP+FN}$ & It is the number of right positive outcomes divided by the number of all related samples (including samples that were meant to be positive).\\   

F1-score & $\frac{2\times P \times R }{P+R}$ & It is the harmonic mean of the precision and recall values.\\   
 
MCC & $\frac{(TP*TN-FP*FN)}{\sqrt{(TP+FP)*(TP+FN)*(TN+FP)*(TN+FN)}}$ & MCC is the correlation coefficient between the actual values of the class and the predicted values of the class. \\   

Specificity & $\frac{TN}{TP+FN}$ & It is used to calculate the fraction of negative values correctly classified. \\

Gmean1 & $\sqrt{Precision \times Recall}$ & Gmean1 is computed as the square root of the product of precision and recall.\\

Gmean2 & $\sqrt{Specificity \times Recall}$ & Gmean2 is computed as the square root of the product of specificity and recall.\\
\hline
\end{tabular}
\vspace{-3.0ex}
\end{table*}

\subsection{Performance Metrics}
In the experimental evaluation, we have used various classification evaluation measures such as Accuracy, Precision, Recall, F1-score, Specificity, G-means, and MCC (Mathews Correlation Coefficient). The reason for choosing the above performance metrics is due to the wide variety of applications of NLP in the current scenario. There are many cases where precision is given more preference over accuracy and similar for other performance metrics. So we evaluated our model TextConvoNet on a multitude of performance metrics in order to get a wholesome analysis of the model and check its viability for each and every NLP application. Table~\ref{PM} shows the descriptions/formulae of all the performance metrics considered.

To assess the statistical significance of the presented \textit{TextConvoNet\_4} and \textit{TextConvoNet\_6} with other considered machine learning and deep learning techniques, we performed the Wilcoxon Signed-Rank paired sample test. It is a non-parametric test, which does not assume the normality of within-pair differences. It tests the hypothesis of whether the median difference is zero between the tested pair or not. We have used a significance level of 95\% (i.e., $\alpha$=0.05) for all the tests. The framed Null Hypothesis ($H_{0}$ and Alternative Hypothesis ($H_{a}$) are as follow.

\noindent \textit{$H_{0}$: No statistically significant variation is there between the paired group for value $\alpha$=0.05.}

\noindent \textit{$H_{a}$:  Statistically, significant variation is present between the paired group for value of $\alpha$=0.05.}

Null hypothesis can be rejected when the experimental p-value has come out to be lesser then the $\alpha$ value, and can be concluded that there is a significant difference between the paired group. If this is not the case then automatically accept the null hypothesis.

Further, we have performed an effect size analysis using Pearson Effect \textit{r} measure. The effect size shows the magnitude of performance difference among the groups. The larger the effect of size the stronger the relationship between two variables. It is defined by Equation \ref{eq:eq11}.

\begin{equation}
r = \frac{z}{\sqrt(2n)}
\label{eq:eq11}
\end{equation}

\noindent Where 2n = number of observations, including the cases where the difference is 0 and \textit{z} is the z-score value defined by Equation \ref{eq:eq12}.

\begin{equation}
z = \frac{|U-\mu|-0.5}{\sigma}
\label{eq:eq12}
\end{equation}

According to Cohen \cite{mcleod2019does}, the effect size is: Low, if r$\approx$0.1; Medium, if r$\approx$0.3; and Large, if r$\approx$0.5.

\subsection{Machine learning and Deep learning techniques used for comparison}
For a comprehensive performance evaluation of the proposed \textit{TextConvoNet}, we have used seven different machine learning techniques namely, Multinomial Naive Bayes~\cite{murphy2006naive}, Decision Tree (DT)~\cite{safavian1991survey}, Random Forest (RF)~\cite{liaw2002classification}, Support Vector Classifier (SVC)~\cite{keerthi2000fast}, Gradient Boosting classifier~\cite{friedman2001greedy}, K-Nearest Neighbour (KNN) ~\cite{zhang2017learning}, and XGBoost~\cite{chen2015xgboost}. A comprehensive evaluation of the proposed \textit{TextConvoNet} using these techniques helps in establishing the usability of the \textit{TextConvoNet} and increases the generalization of the results. Since \textit{TextConvoNet} is a convolutional neural network-based deep learning architecture, we have included some deep learning-based approaches for performance comparison. Specifically, we have implemented Kim's CNN model~\cite{kim2014}, Long Short Term Memory (LSTM)~\cite{hochreiter1997long,adhikari2019rethinking} and VDCNN~\cite{simonyan2014very}  based model proposed for text classification (Table IV) and compared our model with these models. We have also compared our model with other recent attention and/or transformer-based deep learning models such as BERT~\cite{devlin2018bert}, Attention-based BiLSTM~\cite{zhou2016attention}, and Hierarchical attention networks (HAN)~\cite{yang2016hierarchical}. The description and implementation details of these techniques are given as follows. All implementation has been carried out using Python libraries.

\subsubsection{ Kim's CNN model~\cite{kim2014}}  A detailed description of Kim's model has been provided in Section \ref{3(A)}. The implementation details of this model are as follows. In Kim's model, sentences are matched to the embedding vectors that are made available as an input matrix to the model. It uses 3 parallel layers of convolution with word vectors on top obtained from the existing pretrained model, with 100 filters of kernel sizes 3, 4, and 5. It is followed by a dense layer of 64 neurons and a classification layer.

\subsubsection{Long Short Term Memory (LSTM)~\cite{hochreiter1997long,adhikari2019rethinking}} Long Short-Term Memory Networks (LSTMs) is a special form of recurrent neural network (RNN) that can handle long-term dependencies. Also, LSTMs have a chain-like RNN structure, but the repeated module has a different structure. Rather than having a single layer of the neural network, there are four, which communicate in a unique way. Some of the works used the LSTM model for different text classification tasks~\cite{adhikari2019rethinking,yang2018sgm,zhou2016attention,yang2016hierarchical}. For the comparison with our model, we used a single LSTM layer with 32 memory cells in it followed by the classification layer.

\subsubsection{Very Deep Convolutional Neural Networks (VDCNN) \cite{simonyan2014very}} Unlike \textit{TextConvoNet}, which is a shallow network, VDCNN uses multiple layered convolution and max-pooling operation. Therefore, inspired by VDCNN, we implemented its version based on word embedding. This model uses four different pooling processes, each of which reduces the resolution by half, resulting in four different feature map tiers: 64, 128, 256, and 512, followed by a max-pooling layer. After the end of 4 convolution pair operations, the $512 \times k$ resulting features are transformed into a single vector which is the input to a three-layer fully connected classifier (4096,2048,2048) with ReLU hidden units and softmax outputs.  Depending on the classification task, the number of output neurons varies.

\subsubsection{Attention+BiLSTM~\cite{zhou2016attention}}
It starts with an input layer that Tokenize input sentences and indexed lists followed by an embedding layer. There exist bidirectional LSTM cells(100 hidden units) which can be concatenated to get a representation of each token. Attention weights are taken from linear projection and non-linear activation. Final sentence representation is a weighted sum of all token representations. The final classification output is derived from a simple dense and softmax layer.

\subsubsection{BERT~\cite{devlin2018bert}}
The pre-trained BERT model can be fine-tuned with just one additional output layer to create state-of-the-art models for a wide range of NLP tasks, and we have also used the same strategy in order to compare \textit{TextConvoNet} with BERT based model. We used pretrained bert-base-uncased embeddings to encode, followed by a dense layer of 712 neurons ended by a classification layer.

\subsubsection{Hierarchical Attention Networks (HAN)~\cite{yang2016hierarchical}} We set the word embedding dimension to 300 and the GRU dimension to 50 in our experiment. A combination of forward and backward GRU provides us with 100 dimensions for word/sentence annotation in this scenario. The word/sentence context vectors have a dimension of 100 and are randomly initialized. We utilize a 32-piece mini-batch size for training.

\subsection{Implementation Details}
All the experiments to examine the model performance of \textit{TextConvoNet\_4} and \textit{TextConvoNet\_6} is carried out on a system having with Dual-Core Intel Core i5 processor and 8 GB RAM, running Macintosh operating system, with 64- bit processor and access to NVidia K80 GPU kernel. All experiments were performed in Python 3.0. The models are trained over a mini-batch size of 32 using Adam as an optimizer. The Learning rate is chosen to be 0.1, and the models are trained over 10 epochs with Early stopping to avoid overfitting. All these hyperparameters are chosen by making use of a hyperparameter optimization technique called \textit{GridSearchCV}. For generating word vectors from sentences, we use  GloVe\footnote{https://nlp.stanford.edu/projects/glove/} a pre-trained word embedding model.

\begin{table*}[htb]
	\caption{Classification results of the proposed \textit{TextConvoNet} and other models for different performance measures}
	\vspace{-1.0ex}
	\label{binresults}
\scriptsize
	\centering
\begin{tabular}{|p{1.0cm}|p{1.6cm}|p{1.6cm}|p{0.5cm}|p{0.6cm}|p{0.75cm}|p{1.2cm}|p{0.7cm}|p{0.8cm}|p{0.95cm}|p{0.8cm}|p{0.7cm}|p{0.8cm}|}
\hline
\multicolumn{13}{|c|}{\textbf{DATASET-1}}                                                                                                                                                      \\ \hline
            & \textbf{TextConvoNet\_6} & \textbf{TextConvoNet\_4} & \textbf{Kim's Model} & \textbf{LSTM}  & \textbf{VDCNN} & \textbf{Multinomial Naive Bayes} & \textbf{Random Forest} & \textbf{Decision Tree} & \textbf{SVC }  & \textbf{Gradient Boosting} & \textbf{KNN}   & \textbf{XGBoost} \\ \hline
Accuracy    & 0.822           & 0.819           & 0.798       & 0.792 & 0.815 & 0.819                   & 0.741         & 0.649         & 0.775 & 0.689             & 0.577 & 0.736   \\ \hline
Precision   & 0.848           & 0.833           & 0.753       & 0.747 & 0.828 & 0.804                   & 0.718         & 0.646         & 0.756 & 0.646             & 0.599 & 0.711   \\ \hline
Recall      & 0.783           & 0.798           & 0.886       & 0.883 & 0.794 & 0.844                   & 0.792         & 0.656         & 0.811 & 0.832             & 0.462 & 0.794   \\ \hline
F1\_Score   & 0.814           & 0.815           & 0.814       & 0.809 & 0.811 & 0.823                   & 0.753         & 0.651         & 0.782 & 0.727             & 0.522 & 0.75    \\ \hline
Specificity & 0.86            & 0.841           & 0.711       & 0.702 & 0.836 & 0.795                   & 0.69          & 0.641         & 0.739 & 0.546             & 0.692 & 0.678   \\ \hline
Gmean1      & 0.815           & 0.815           & 0.817       & 0.812 & 0.811 & 0.824                   & 0.754         & 0.651         & 0.783 & 0.733             & 0.526 & 0.751   \\ \hline
Gmean2      & 0.821           & 0.819           & 0.793       & 0.787 & 0.815 & 0.819                   & 0.739         & 0.649         & 0.774 & 0.674             & 0.565 & 0.734   \\ \hline
MCC         & 0.645           & 0.639           & 0.605       & 0.595 & 0.63  & 0.639                   & 0.484         & 0.297         & 0.551 & 0.394             & 0.158 & 0.475   \\ \hline
\multicolumn{13}{|c|}{\textbf{DATASET-2}}                                                                                                                                                      \\ \hline
            & \textbf{TextConvoNet\_6} & \textbf{TextConvoNet\_4} & \textbf{Kim's Model} & \textbf{LSTM}  & \textbf{VDCNN} & \textbf{Multinomial Naive Bayes} & \textbf{Random Forest} & \textbf{Decision Tree} & \textbf{SVC }  & \textbf{Gradient Boosting} & \textbf{KNN}   & \textbf{XGBoost} \\ \hline
Accuracy    & 0.909           & 0.871           & 0.885       & 0.877 & 0.77  & 0.824                   & 0.75          & 0.689         & 0.781 & 0.786             & 0.723 & 0.809   \\ \hline
Precision   & 0.909           & 0.829           & 0.867       & 0.849 & 0.968 & 0.831                   & 0.764         & 0.774         & 0.788 & 0.799             & 0.737 & 0.838   \\ \hline
Recall      & 0.862           & 0.905           & 0.859       & 0.83  & 0.467 & 0.921                   & 0.976         & 0.784         & 0.967 & 0.928             & 0.956 & 0.908   \\ \hline
F1\_Score   & 0.89            & 0.855           & 0.861       & 0.845 & 0.61  & 0.812                   & 0.849         & 0.779         & 0.868 & 0.858             & 0.813 & 0.845   \\ \hline
Specificity & 0.939           & 0.859           & 0.888       & 0.904 & 0.983 & 0.556                   & 0.259         & 0.451         & 0.361 & 0.442             & 0.178 & 0.576   \\ \hline
Gmean1      & 0.876           & 0.861           & 0.843       & 0.852 & 0.694 & 0.881                   & 0.859         & 0.781         & 0.867 & 0.857             & 0.839 & 0.868   \\ \hline
Gmean2      & 0.901           & 0.879           & 0.879       & 0.87  & 0.686 & 0.715                   & 0.476         & 0.582         & 0.58  & 0.651             & 0.43  & 0.711   \\ \hline
MCC         & 0.807           & 0.741           & 0.75        & 0.75  & 0.565 & 0.532                   & 0.334         & 0.229         & 0.428 & 0.428             & 0.20  & 0.51    \\ \hline
\multicolumn{13}{|c|}{\textbf{DATASET-3}}                                                                                                                                                      \\ \hline
            & \textbf{TextConvoNet\_6} & \textbf{TextConvoNet\_4} & \textbf{Kim's Model} & \textbf{LSTM}  & \textbf{VDCNN} & \textbf{Multinomial Naive Bayes} & \textbf{Random Forest} & \textbf{Decision Tree} & \textbf{SVC }  & \textbf{Gradient Boosting} & \textbf{KNN}   & \textbf{XGBoost} \\ \hline
Accuracy    & 0.992           & 0.992           & 0.986       & 0.874 & 0.965 & 0.99                    & 0.983         & 0.977         & 0.978 & 0.986             & 0.972 & 0.99    \\ \hline
Precision   & 0.968           & 0.969           & 0.945       & 0.495 & 0.862 & 0.958                   & 0.931         & 0.91          & 0.914 & 0.944             & 0.887 & 0.96    \\ \hline
Recall      & 0.968           & 0.969           & 0.945       & 0.495 & 0.862 & 0.958                   & 0.931         & 0.91          & 0.914 & 0.944             & 0.887 & 0.96    \\ \hline
F1\_Score   & 0.968           & 0.969           & 0.945       & 0.495 & 0.862 & 0.958                   & 0.931         & 0.91          & 0.914 & 0.944             & 0.887 & 0.96    \\ \hline
Specificity & 0.995           & 0.996           & 0.992       & 0.928 & 0.98  & 0.994                   & 0.99          & 0.987         & 0.988 & 0.992             & 0.984 & 0.994   \\ \hline
Gmean1      & 0.968           & 0.969           & 0.945       & 0.495 & 0.862 & 0.958                   & 0.931         & 0.91          & 0.914 & 0.944             & 0.887 & 0.96    \\ \hline
Gmean2      & 0.981           & 0.982           & 0.968       & 0.678 & 0.919 & 0.976                   & 0.96          & 0.948         & 0.95  & 0.968             & 0.934 & 0.977   \\ \hline
MCC         & 0.963           & 0.965           & 0.937       & 0.423 & 0.842 & 0.952                   & 0.921         & 0.897         & 0.901 & 0.936             & 0.871 & 0.955   \\ \hline
\multicolumn{13}{|c|}{\textbf{DATASET-4}}                                                                                                                                                      \\ \hline
            & \textbf{TextConvoNet\_6} & \textbf{TextConvoNet\_4} & \textbf{Kim's Model} & \textbf{LSTM}  & \textbf{VDCNN} & \textbf{Multinomial Naive Bayes} & \textbf{Random Forest} & \textbf{Decision Tree} & \textbf{SVC }  & \textbf{Gradient Boosting} & \textbf{KNN}   & \textbf{XGBoost} \\ \hline
Accuracy    & 0.884           & 0.873           & 0.869       & 0.876 & 0.871 & 0.87                    & 0.872         & 0.807         & 0.881 & 0.866             & 0.662 & 0.877   \\ \hline
Precision   & 0.826           & 0.809           & 0.803       & 0.813 & 0.806 & 0.804                   & 0.809         & 0.71          & 0.821 & 0.8               & 0.493 & 0.815   \\ \hline
Recall      & 0.826           & 0.809           & 0.803       & 0.813 & 0.806 & 0.804                   & 0.809         & 0.71          & 0.821 & 0.8               & 0.493 & 0.815   \\ \hline
F1\_Score   & 0.826           & 0.809           & 0.803       & 0.813 & 0.806 & 0.804                   & 0.809         & 0.71          & 0.821 & 0.8               & 0.493 & 0.815   \\ \hline
Specificity & 0.913           & 0.904           & 0.901       & 0.907 & 0.903 & 0.902                   & 0.904         & 0.855         & 0.911 & 0.9               & 0.746 & 0.908   \\ \hline
Gmean1      & 0.826           & 0.809           & 0.803       & 0.813 & 0.806 & 0.804                   & 0.809         & 0.71          & 0.821 & 0.8               & 0.493 & 0.815   \\ \hline
Gmean2      & 0.869           & 0.855           & 0.851       & 0.859 & 0.853 & 0.852                   & 0.855         & 0.779         & 0.865 & 0.848             & 0.606 & 0.86    \\ \hline
MCC         & 0.739           & 0.713           & 0.704       & 0.72  & 0.709 & 0.706                   & 0.713         & 0.565         & 0.732 & 0.699             & 0.239 & 0.723   \\ \hline
\multicolumn{13}{|c|}{\textbf{DATASET-5}}                                                                                                                                                      \\ \hline
            & \textbf{TextConvoNet\_6} & \textbf{TextConvoNet\_4} & \textbf{Kim's Model} & \textbf{LSTM}  & \textbf{VDCNN} & \textbf{Multinomial Naive Bayes} & \textbf{Random Forest} & \textbf{Decision Tree} & \textbf{SVC }  & \textbf{Gradient Boosting} & \textbf{KNN}   & \textbf{XGBoost} \\ \hline
Accuracy    & 0.839           & 0.828           & 0.804       & 0.702 & 0.762 & 0.754                   & 0.75          & 0.729         & 0.76  & 0.78              & 0.69  & 0.785   \\ \hline
Precision   & 0.597           & 0.571           & 0.51        & 0.255 & 0.404 & 0.385                   & 0.375         & 0.321         & 0.401 & 0.449             & 0.224 & 0.461   \\ \hline
Recall      & 0.597           & 0.571           & 0.51        & 0.255 & 0.404 & 0.385                   & 0.375         & 0.321         & 0.401 & 0.449             & 0.224 & 0.461   \\ \hline
F1\_Score   & 0.597           & 0.571           & 0.51        & 0.255 & 0.404 & 0.385                   & 0.375         & 0.321         & 0.401 & 0.449             & 0.224 & 0.461   \\ \hline
Specificity & 0.899           & 0.893           & 0.878       & 0.814 & 0.851 & 0.846                   & 0.844         & 0.83          & 0.85  & 0.862             & 0.806 & 0.865   \\ \hline
Gmean1      & 0.597           & 0.571           & 0.51        & 0.255 & 0.404 & 0.385                   & 0.375         & 0.321         & 0.401 & 0.449             & 0.224 & 0.461   \\ \hline
Gmean2      & 0.732           & 0.714           & 0.669       & 0.456 & 0.586 & 0.571                   & 0.562         & 0.517         & 0.584 & 0.622             & 0.425 & 0.632   \\ \hline
MCC         & 0.496           & 0.463           & 0.388       & 0.069 & 0.255 & 0.232                   & 0.218         & 0.152         & 0.251 & 0.312             & 0.03  & 0.327   \\ \hline
\end{tabular}
\end{table*}

\subsection{Results and Discussion}\label{results}
In the first experiment, we tested the performance of \textit{TextConvoNet} on five datasets (Section \ref{IV-A}). Table~\ref{binresults} shows the performance of \textit{TextConvoNet} (\textit{TextConvoNet\_6} and  \textit{TextConvoNet\_4}) in comparison with baseline machine learning models and state-of-the-art Deep Learning based models. In Table \ref{binresults}, results are reported for binary classification datasets (2 and 4) and multi-class classification datasets (3, 4 and 5). From the results, following inferences can be drawn.

\begin{itemize}
    \item In terms of accuracy, all the deep learning models are producing value above than 70\%. The highest value is for dataset-3 and it is given by the proposed \textit{TextConvoNet\_6}. The minimum average gain in accuracy\ over all datasets is reported to be 1.5\% with maximum average gain of 31\%. 
    \item In terms of precision, \textit{TextConvoNet} performs better than all other models for all datasets except dataset-2 with minimum average gain in precision over all datasets to be 4.1\% and maximum average gain of 70.8\% with almost similar results for dataset-2. 
    \item In terms of recall, \textit{TextConvoNet} produces almost comparable results for dataset-1 and dataset-2 with a minimum average gain of 3.7\% and a maximum average gain of 65.8\% alongside the best results on dataset-5.
    \item The average gain values for F1\_Score was also found to similar to the Precision with \textit{TextConvoNet} performing best on dataset-5.
    \item There was a minimum average gain of 1.1\% and a maximum average gain of  18.6\% in terms of Specificity. \textit{TextConvoNet} produced almost comparable results for dataset-2 than other machine and deep learning models.
    \item The minimum and maximum average gain in Gmean1 and Gmean2 is 3.9\%,68\% and 2.7\%,65.66\% respectively.
    \item In terms of MCC, \textit{TextConvoNet} performs extremely well on all the datasets with minimum average gain of 7.4\% and a maximum average gain of about 30\%.
    \item Overall, in multi-class datasets (3, 4 and 5), variants of \textit{TextConvoNet} perform better than all the other models in terms of all the performance metrics.
 \end{itemize}

\begin{table*}[htb]
	\centering
	\scriptsize
	\caption{Result comparison of \textit{TextConvoNet} with attention and/or transformer-based deep learning models}
	\vspace{-1.5ex}
	\begin{tabular}{l|l|l|l|l|l|l|l|l}
\hline
\multicolumn{9}{c}{\textbf{DATASET-1}}                                                                     \\ \hline
                   & Accuracy  & Precision & Recall  & F1\_score & Specificity & Gmean 1 & Gmean 2 & MCC     \\ \hline
BiLSTM + Attention & 0.8066    & 0.816     & 0.7909  & 0.803     & 0.8223      & 0.803   & 0.806   & 0.6136  \\ \hline
BERT               & 0.7764    & 0.7367    & 0.859   & 0.7932    & 0.694       & 0.7956  & 0.772   & 0.56    \\ \hline
HAN                & 0.8028    & 0.7913    & 0.8217  & 0.80626   & 0.7839      & 0.806   & 0.802   & 0.606   \\ \hline
TextConvoNet\_6    & 0.822     & 0.848     & 0.783   & 0.814     & 0.86        & 0.815   & 0.821   & 0.645   \\ \hline
TextConvoNet\_4    & 0.819     & 0.833     & 0.798   & 0.815     & 0.841       & 0.815   & 0.819   & 0.639   \\ \hline
\multicolumn{9}{c}{\textbf{DATASET-2}}                                                                     \\ \hline
BiLSTM + Attention & 0.886     & 0.8834    & 0.8465  & 0.864     & 0.9157      & 0.8648  & 0.8804  & 0.7667  \\ \hline
BERT               & 0.772     & 0.6856    & 0.8674  & 0.76591   & 0.7         & 0.7712  & 0.7792  & 0.564   \\ \hline
HAN                & 0.863     & 0.8805    & 0.7883  & 0.8319    & 0.9192      & 0.8331  & 0.8513  & 0.72    \\ \hline
TextConvoNet\_6    & 0.904     & 0.905     & 0.867   & 0.886     & 0.932       & 0.886   & 0.899   & 0.804   \\ \hline
TextConvoNet\_4    & 0.872     & 0.819     & 0.902   & 0.858     & 0.849       & 0.859   & 0.875   & 0.745   \\ \hline
\multicolumn{9}{c}{\textbf{DATASET-3}}                                                                     \\ \hline
BiLSTM + Attention & 0.99      & 0.96      & 0.9602  & 0.9602    & 0.99432     & 0.9602  & 0.9771  & 0.9545  \\ \hline
BERT               & 0.956487  & 0.8259    & 0.82594 & 0.82594   & 0.97513     & 0.82594 & 0.8974  & 0.801   \\ \hline
HAN                & 0.8901325 & 0.5605    & 0.5605  & 0.56052   & 0.9372      & 0.5605  & 0.7248  & 0.4977  \\ \hline
TextConvoNet\_6    & 0.992     & 0.968     & 0.968   & 0.968     & 0.995       & 0.968   & 0.981   & 0.963   \\ \hline
TextConvoNet\_4    & 0.992     & 0.969     & 0.969   & 0.969     & 0.996       & 0.969   & 0.982   & 0.965   \\ \hline
\multicolumn{9}{c}{\textbf{DATASET-4}}                                                                     \\ \hline
BiLSTM + Attention & 0.8619253 & 0.7928    & 0.7928  & 0.79288   & 0.896       & 0.7928  & 0.843   & 0.68933 \\ \hline
BERT               & 0.808     & 0.712     & 0.712   & 0.712     & 0.856       & 0.712   & 0.7806  & 0.568   \\ \hline
HAN                & 0.7955556 & 0.69333   & 0.6933  & 0.69333   & 0.84666     & 0.6933  & 0.7661  & 0.54    \\ \hline
TextConvoNet\_6    & 0.884     & 0.826     & 0.826   & 0.826     & 0.913       & 0.826   & 0.869   & 0.739   \\ \hline
TextConvoNet\_4    & 0.873     & 0.809     & 0.809   & 0.809     & 0.904       & 0.0.809 & 0.855   & 0.713   \\ \hline
\multicolumn{9}{c}{\textbf{DATASET-5}}                                                                     \\ \hline
BiLSTM + Attention & 0.7114667 & 0.2786    & 0.2786  & 0.27866   & 0.81966     & 0.2786  & 0.4779  & 0.0983  \\ \hline
BERT               & 0.7114667 & 0.2786    & 0.27866 & 0.27866   & 0.81966     & 0.27866 & 0.47792 & 0.0983  \\ \hline
HAN                & 0.7       & 0.25      & 0.25    & 0.25      & 0.8125      & 0.25    & 0.45069 & 0.0625  \\ \hline
TextConvoNet\_6    & 0.839     & 0.597     & 0.597   & 0.597     & 0.899       & 0.597   & 0.732   & 0.496   \\ \hline
TextConvoNet\_4    & 0.828     & 0.571     & 0.571   & 0.571     & 0.893       & 0.571   & 0.714   & 0.463   \\ \hline
\end{tabular}
	\label{f1}
\end{table*}

\begin{table*}[thb]
\centering
\scriptsize
\caption{Results of Wilcoxon signed-rank paired sampled test between the presented \textit{TextConvoNet\_6} and other techniques}
\vspace{-1.0ex}
\label{prvaluetext6}
\begin{tabular}{lllllp{1.3cm}p{1.0cm}p{1.0cm}lp{1.0cm}lp{1.0cm}}
\hline
\multicolumn{12}{c}{\textbf{\textit{TextConvoNet\_6} (Two-tailed)}}                                                                                                                                                   \\ \hline
             & Textconvonet\_4 & Kim's CNN  & LSTM     & VDCNN    & Multinomial Naïve Bayes & Random Forest & Decision Tree & SVC      & Gradient Boosting Classifier & KNN      & XGBoost \\ \hline
\textit{P}-value      & 3.36E-05        & 1.63E-07 & 4.66E-08 & 2.77E-07  & 1.72E-06                & 6.4E-08       & 6.5E-09       & 1.76E-07 & 1.76E-07                     & 1.07E-08 & 1.2E-07 \\
Effect \textit{r}   & 0.463           & 0.585    & 0.61     & 0.574     & 0.534                   & 0.604         & 0.648         & 0.83     & 0.583                        & 0.639    & 0.591   \\
Sig.   Diff. & Yes             & Yes      & Yes      & Yes        & Yes                     & Yes           & Yes           & Yes      & Yes                          & Yes      & Yes    \\ \hline
\end{tabular}
\vspace{-1.0ex}
\end{table*}

\begin{table*}[thb]
\centering
\scriptsize
\caption{Results of Wilcoxon signed-rank paired sampled test between the presented \textit{TextConvoNet\_4} and other techniques}
\vspace{-1.0ex}
\label{prvaluetext4}
\begin{tabular}{llllp{1.3cm}p{1.0cm}p{1.0cm}lp{1.6cm}lp{1.0cm}}
\hline
\multicolumn{11}{c}{\textbf{\textit{TextConvoNet\_4} (Two-tailed)}}                                                                                                                                 \\ \hline
             & Kim's CNN & LSTM     & VDCNN      & Multinomial Naïve Bayes & Random Forest & Decision Tree & SVC      & Gradient Boosting Classifier & KNN      & XGBoost  \\ \hline
\textit{P}-value      & 0.00017 & 3.36E-05 & 6.75E-07  & 0.00039                 & 4.66E-08      & 6.5E-09       & 1.22E-05 & 1.29E-07                     & 9.89E-09 & 5.82E-05 \\
Effect \textit{r}   & 0.419   & 0.463    & 0.55     & 0.395                   & 0.610         & 0.648         & 0.489    & 0.590                        & 0.640    & 0.449    \\
Sig. Diff. & Yes     & Yes        & Yes      & Yes                     & Yes           & Yes           & Yes      & Yes                          & Yes      & Yes     \\ \hline
\end{tabular}
\vspace{-3.0ex}
\end{table*}

We have also compared the results of \textit{TextConvoNet} with two attention-based model: BiLSTM followed by attention (Attention+BiLSTM) \& Hierarchical Attention Network (HAN) and one transformer-based BERT. The  Table \ref{f1} shows the results of  \textit{TextConvoNet} with BiLSTM +Attention, BERT and HAN on different performance measures. From the table, we can observe, the HAN has poor results on the all the datasets in comparison with other models. On dataset 1 and 3, \textit{TextConvoNet\_4} has the better results in comparison with others. On dataset 2, 4, and 5  \textit{TextConvoNet\_6} outperforms others. Overall, the results of \textit{TextConvoNet} are better in comparison with other models.'

\subsubsection{Wilcoxon signed-rank test}
Tables \ref{prvaluetext6} and \ref{prvaluetext4} report the Wilcoxon signed-rank test results in terms of p-values and effect \textit{r} for the \textit{TextConvoNet\_6} and \textit{TextConvoNet\_4} with other techniques, respectively. The last row in the tables shows whether a statistically significant performance difference has been found between the compared groups. For Table \ref{prvaluetext6}, it is observed that there is a statistically significant difference between the presented \textit{TextConvoNet\_6} and all other considered techniques. The experimental p-values are less than the significance level of 0.05 in all groups. Further, effect \textit{r} values are higher than 0.45 in all the groups, showing a large magnitude of performance difference between the \textit{TextConvoNet\_6} and other techniques. Similarly, from Table \ref{prvaluetext4}, it is observed that the performance difference between the presented \textit{TextConvoNet\_4} and other techniques is statistically significant at the given significance level for all cases.
A significant difference can be seen in all the groups as the P-values are below 0.05. Further, effect \textit{r} values are higher than 0.40 for all the groups showing a large magnitude of the performance difference.

\begin{table}[htb]

\scriptsize
\centering
\caption{Different versions of \textit{TextConvoNet}}
\label{Version}
\begin{tabular}{p{0.3cm}p{0.85cm}p{0.7cm}p{0.7cm}p{0.7cm}p{1.5cm}p{1.5cm}}
\hline
     & \# Filters & Dropout Rate & Optimizers & Dense Layer (units) & Kernel Size (intra-sentence layers) & Kernel sizes (inter-sentence layers) \\ \hline
V1.1 & 32                & 0.4          & adam       & 64                 & (1X2,1X3,1X4)                     & (2X1,2X2,2X3)                       \\
V1.2 & 32                & 0.4          & RMSProp    & 64                 & (1X2,1X3,1X4)                     & (2X1,2X2,2X3)                       \\
V1.3 & 48                & 0.4          & adam       & 64                 & (1X2,1X3,1X4)                     & (2X1,2X2,2X3)                       \\
V1.4 & 48                & 0.5          & adam       & 64                 & (1X2,1X3,1X4)                     & (2X1,2X2,2X3)                       \\
V1.5 & 32                & 0.4          & adam       & 96                 & (1X2,1X3,1X4)                     & (2X1,2X2,2X3)                       \\
V1.6 & 32                & 0.5          & adam       & 96                 & (1X2,1X3,1X4)                     & (2X1,2X2,2X3)                       \\ \hdashline
     &                   &              &            &                    &                                   &                                     \\ 
V2.1 & 32                & 0.4          & adam       & 64                 & (1X3,1X4,1X5)                     & (2X2,2X3,2X4)                        \\
V2.2 & 32                & 0.4          & RMSProp    & 64                 & (1X3,1X4,1X5)                     & (2X2,2X3,2X4)                        \\
V2.3 & 48                & 0.4          & adam       & 64                 & (1X3,1X4,1X5)                     & (2X2,2X3,2X4)                        \\
V2.4 & 48                & 0.5          & adam       & 64                 & (1X3,1X4,1X5)                     & (2X2,2X3,2X4)                        \\
V2.5 & 32                & 0.4          & adam       & 96                 & (1X3,1X4,1X5)                     & (2X2,2X3,2X4)                        \\
V2.6 & 32                & 0.5          & adam       & 96                 & (1X3,1X4,1X5)                     & (2X2,2X3,2X4)                        \\ \hdashline
    &                   &              &            &                    &                                   &                                     \\
V3.1 & 32                & 0.4          & adam       & 64                 & (1X2,1X3,1X4)                     & (3X1,3X2,3X3)                       \\
V3.2 & 32                & 0.4          & RMSProp    & 64                 & (1X2,1X3,1X4)                     & (3X1,3X2,3X3)                       \\
V3.3 & 48                & 0.4          & adam       & 64                 & (1X2,1X3,1X4)                     & (3X1,3X2,3X3)                       \\
V3.4 & 48                & 0.5          & adam       & 64                 & (1X2,1X3,1X4)                     & (3X1,3X2,3X3)                       \\
V3.5 & 32                & 0.4          & adam       & 96                 & (1X2,1X3,1X4)                     & (3X1,3X2,3X3)                       \\
V3.6 & 32                & 0.5          & adam       & 96                 & (1X2,1X3,1X4)                     & (3X1,3X2,3X3)                       \\ \hdashline
     &                   &              &            &                    &                                   &                                     \\
V4.1 & 32                & 0.4          & adam       & 64                 & (1X3,1X4,1X5)                     & (3X2,3X3,3X4)                       \\
V4.2 & 32                & 0.4          & RMSProp    & 64                 & (1X3,1X4,1X5)                     & (3X2,3X3,3X4)                       \\
V4.3 & 48                & 0.4          & adam       & 64                 & (1X3,1X4,1X5)                     & (3X2,3X3,3X4)                       \\
V4.4 & 48                & 0.5          & adam       & 64                 & (1X3,1X4,1X5)                     & (3X2,3X3,3X4)                       \\
V4.5 & 32                & 0.4          & adam       & 96                 & (1X3,1X4,1X5)                     & (3X2,3X3,3X4)                       \\
V4.6 & 32                & 0.5          & adam       & 96                 & (1X3,1X4,1X5)                     & (3X2,3X3,3X4)     \\ \hline                 
\end{tabular}
\vspace{-3.0ex}
\end{table}

\begin{figure*}[htb]
  \centering
    \begin{subfigure}{0.40\textwidth}
  \includegraphics[width=3.0in,height=2.1in,angle=0]{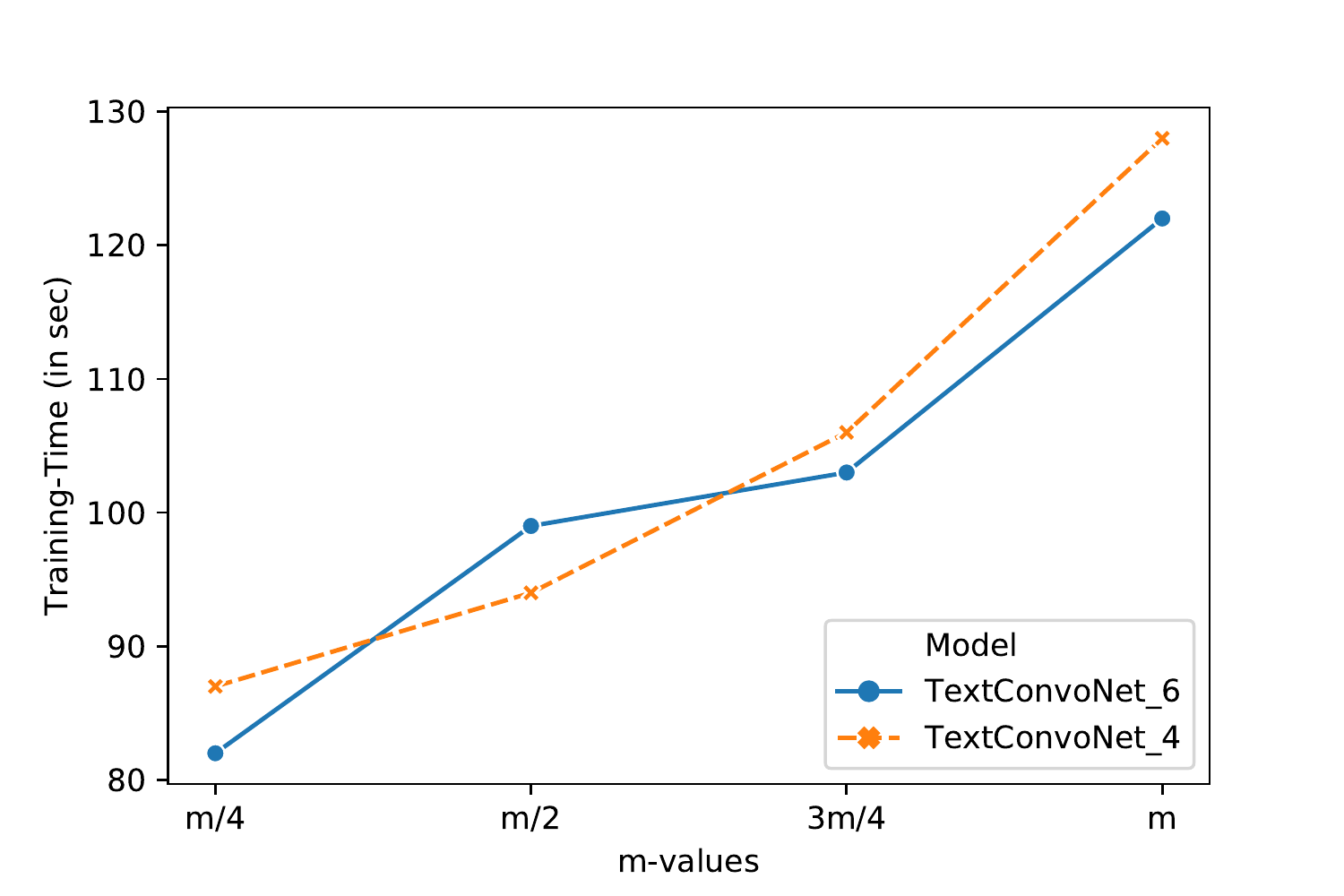}\quad
   \caption{\textit{m} on Dataset 2}
  \label{fig:1}
\end{subfigure}%
    \begin{subfigure}{0.40\textwidth}
  \includegraphics[width=3.0in,height=2.1in,angle=0]{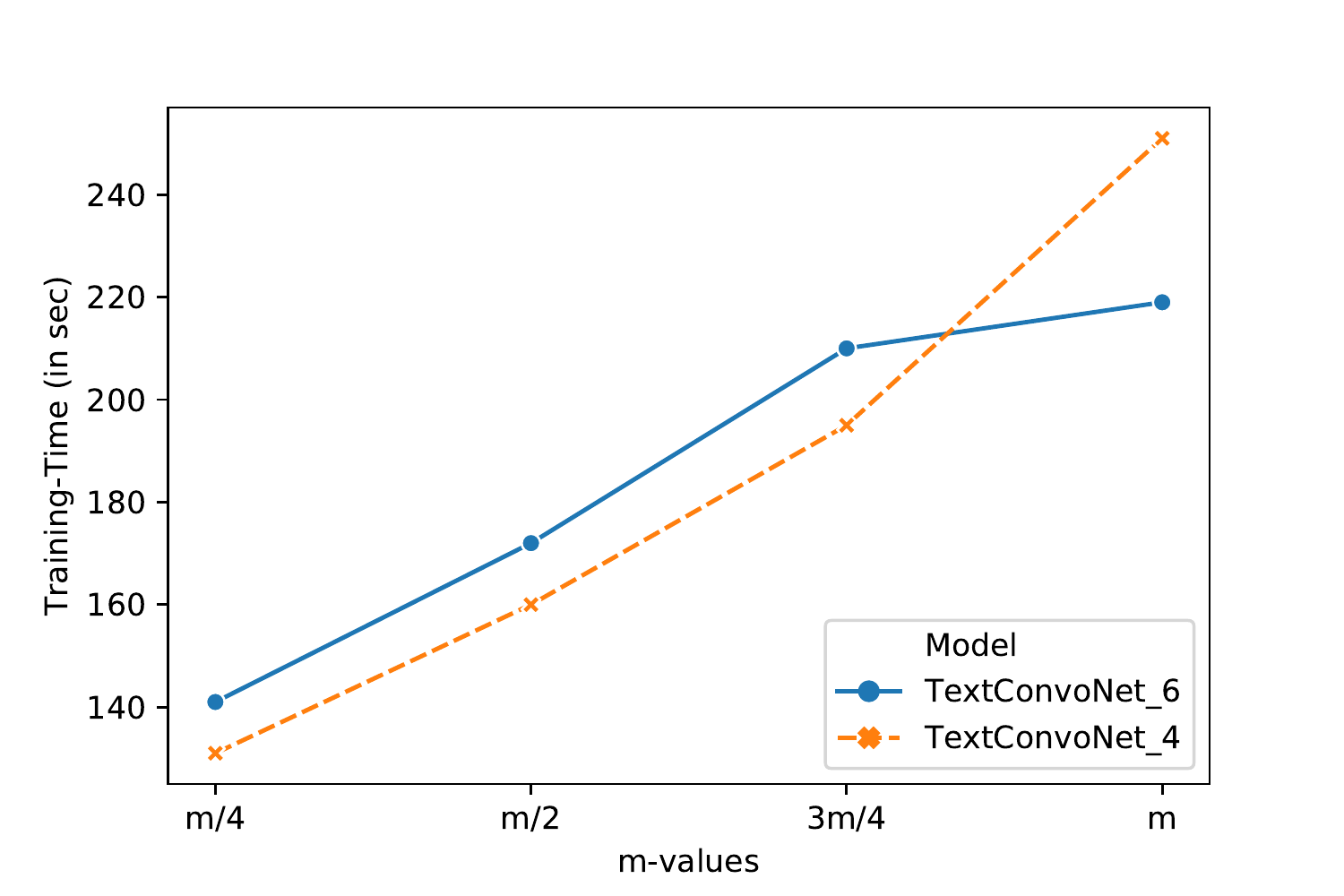}\quad
   \caption{\textit{m} on Dataset 5}
  \label{fig:1}
\end{subfigure}%

  \caption{Training Time with different values of \textit{m}}
  \label{runtime}
  \vspace{-2.0ex}
\end{figure*}

\subsection{Ablation Study}
We evaluated our model variant \textit{TextConvoNet\_6} over a variety of parameters given in Table \ref{Version} to analyse their effect on the various performance metrics. Between any two versions, there is a change in the kernel size. Within a version, between any two sub-versions, there are changes in the number of filters, dropout rate before the classification layer, optimizer, and the units in the fully connected layer. Generally, performance metrics can change adhering to the needs and application and the type of model. In practice, each NLP application is unique. Therefore, for every application of NLP, a unique approach/model is needed. Hence, we have performed an ablation study on the selected Datasets (Table \ref{datasetdetails}). The results are shown in Table \ref{ablation}.

From the results between various versions of \textit{TextConvoNet}, the following observations have been drawn. 

\begin{itemize}
    \item We conclude that Adam is the best optimizer for the model as other optimizers were taking a large amount of time train on or did not give good results as in the case of RMSProp (V1.2, V2.2, V3.2, V4.2).
    \item Dropout Rate of 0.4 was found to be optimum as for the value 0.5 model was slightly overfitting.
    \item In datasets with longer texts (like Dataset-2 and Dataset-5)
    Versions 3 and 4 seem to give slightly better results when compared to Version 1 and 2. The large kernel sizes in Versions 3 and 4 might be the possible reason behind the inference.
    \item Whereas datasets (Datasets 1, 3, and 4) with comparatively smaller paragraphs do well in Versions 1 and 2.
\end{itemize}

\noindent\textbf{Optimum value of \textit{m}:} To evaluate our models' run-time performance, we tracked down the training time by varying the value of \textit{(m)}. It is quite evident that varying the value of \textit{m} ranging from ($\frac{m}{4}$ to \textit{m}) also leads to an increase in the computation time. The results for run-time performance of \textit{TextConvoNet} are shown in Fig \ref{runtime}. The value of \textit{m} was also varied from $\frac{m}{4}$ to $m$ and the performance metrics for each of the above m-values was calculated. The results are mentioned in the Table \ref{mtable}. The observations obtained from the results are as follow.

\begin{itemize}
    \item It was observed that the performance of \textit{TextConvoNet\_6} slightly betters itself in all the metrics while increasing the value from $\frac{m}{4}$ to $m$ on datasets having a maximum length of sentences in a paragraph considerably small, as shown for (Dataset-4).
    \item Datasets having maximum length of sentences in a paragraph considerably larger, \textit{TextConvoNet\_6} is seen to perform well on lower values of $m(\frac{m}{4},\frac{m}{2},\frac{3m}{4})$. 
   
\end{itemize}
 
The possible reason for the above observations could be that in smaller paragraphs, \textit{TextConvoNet\_6} finds it difficult to extract the features from the text due to lesser data and hence require a large number of sentences to work on, as evident in (Dataset-4). On the other hand \textit{TextConvoNet\_6} won’t mind dropping a portion of sentences like in (Dataset-2) having larger paragraphs due to ample amount of textual data already present.

\begin{table*}[htb]
\centering

\tiny
\caption{Results of accuracy, precision, and recall evaluation measures on different datasets}
\label{ablation}
\begin{tabular}{lp{0.7cm}p{0.7cm}p{0.7cm}p{0.7cm}p{0.7cm}|p{0.7cm}p{0.7cm}p{0.7cm}p{0.7cm}p{0.7cm}|p{0.7cm}p{0.7cm}p{0.7cm}p{0.7cm}p{0.7cm}}
\hline
\multicolumn{6}{c}{\textbf{Accuracy}}                            & \multicolumn{5}{c}{\textbf{Precision}}                    & \multicolumn{5}{c}{\textbf{Recall}}                       \\ \hline
     & Dataset-1 & Dataset-2 & Dataset-3 & Dataset-4 & Dataset-5 & Dataset-1 & Dataset-2 & Dataset-3 & Dataset-4 & Dataset-5 & Dataset-1 & Dataset-2 & Dataset-3 & Dataset-4 & Dataset-5 \\ \hline
V1.1 & 0.822     & 0.904     & 0.992     & 0.884     & 0.839     & 0.848     & 0.905     & 0.968     & 0.826     & 0.597     & 0.783     & 0.867     & 0.968     & 0.826     & 0.597     \\
V1.2 & 0.825     & 0.880     & 0.991     & 0.867     & 0.799     & 0.811     & 0.891     & 0.963     & 0.801     & 0.497     & 0.848     & 0.821     & 0.963     & 0.801     & 0.497     \\
V1.3 & 0.820     & 0.890     & 0.992     & 0.874     & 0.835     & 0.822     & 0.900     & 0.968     & 0.811     & 0.587     & 0.816     & 0.837     & 0.968     & 0.811     & 0.587     \\
V1.4 & 0.821     & 0.893     & 0.992     & 0.883     & 0.836     & 0.806     & 0.919     & 0.970     & 0.825     & 0.589     & 0.845     & 0.823     & 0.970     & 0.825     & 0.589     \\
V1.5 & 0.823     & 0.890     & 0.992     & 0.880     & 0.824     & 0.802     & 0.864     & 0.968     & 0.821     & 0.561     & 0.856     & 0.884     & 0.968     & 0.821     & 0.561     \\
V1.6 & 0.805     & 0.896     & 0.992     & 0.868     & 0.825     & 0.762     & 0.916     & 0.968     & 0.802     & 0.563     & 0.886     & 0.835     & 0.968     & 0.802     & 0.563     \\
V2.1 & 0.823     & 0.880     & 0.992     & 0.880     & 0.818     & 0.806     & 0.846     & 0.969     & 0.820     & 0.545     & 0.850     & 0.881     & 0.969     & 0.820     & 0.545     \\
V2.2 & 0.807     & 0.842     & 0.992     & 0.876     & 0.806     & 0.775     & 0.942     & 0.970     & 0.815     & 0.515     & 0.865     & 0.674     & 0.970     & 0.815     & 0.515     \\
V2.3 & 0.825     & 0.891     & 0.991     & 0.879     & 0.829     & 0.801     & 0.853     & 0.964     & 0.819     & 0.573     & 0.864     & 0.902     & 0.964     & 0.819     & 0.573     \\
V2.4 & 0.825     & 0.896     & 0.991     & 0.886     & 0.831     & 0.821     & 0.916     & 0.963     & 0.830     & 0.579     & 0.829     & 0.835     & 0.963     & 0.830     & 0.579     \\
V2.5 & 0.826     & 0.890     & 0.991     & 0.875     & 0.827     & 0.805     & 0.883     & 0.966     & 0.812     & 0.567     & 0.860     & 0.858     & 0.966     & 0.812     & 0.567     \\
V2.6 & 0.811     & 0.892     & 0.994     & 0.892     & 0.828     & 0.764     & 0.874     & 0.978     & 0.838     & 0.569     & 0.900     & 0.874     & 0.978     & 0.838     & 0.569     \\
V3.1 & 0.826     & 0.884     & 0.992     & 0.879     & 0.824     & 0.825     & 0.881     & 0.968     & 0.819     & 0.561     & 0.826     & 0.844     & 0.968     & 0.819     & 0.561     \\
V3.2 & 0.794     & 0.874     & 0.992     & 0.854     & 0.806     & 0.732     & 0.817     & 0.966     & 0.782     & 0.516     & 0.925     & 0.912     & 0.966     & 0.782     & 0.516     \\
V3.3 & 0.826     & 0.894     & 0.994     & 0.882     & 0.832     & 0.805     & 0.870     & 0.976     & 0.823     & 0.580     & 0.861     & 0.886     & 0.976     & 0.823     & 0.580     \\
V3.4 & 0.825     & 0.878     & 0.994     & 0.880     & 0.827     & 0.826     & 0.822     & 0.974     & 0.820     & 0.567     & 0.822     & 0.914     & 0.974     & 0.820     & 0.567     \\
V3.5 & 0.820     & 0.894     & 0.992     & 0.884     & 0.822     & 0.799     & 0.911     & 0.967     & 0.826     & 0.554     & 0.855     & 0.835     & 0.967     & 0.826     & 0.554     \\
V3.6 & 0.818     & 0.886     & 0.992     & 0.869     & 0.824     & 0.785     & 0.907     & 0.968     & 0.804     & 0.560     & 0.875     & 0.819     & 0.968     & 0.804     & 0.560     \\
V4.1 & 0.829     & 0.903     & 0.992     & 0.878     & 0.825     & 0.808     & 0.917     & 0.968     & 0.817     & 0.563     & 0.861     & 0.851     & 0.968     & 0.817     & 0.563     \\
V4.2 & 0.800     & 0.851     & 0.991     & 0.871     & 0.790     & 0.885     & 0.932     & 0.962     & 0.806     & 0.476     & 0.688     & 0.705     & 0.962     & 0.806     & 0.476     \\
V4.3 & 0.812     & 0.897     & 0.992     & 0.878     & 0.826     & 0.783     & 0.885     & 0.968     & 0.817     & 0.564     & 0.862     & 0.874     & 0.968     & 0.817     & 0.564     \\
V4.4 & 0.822     & 0.889     & 0.990     & 0.885     & 0.827     & 0.804     & 0.867     & 0.962     & 0.828     & 0.567     & 0.850     & 0.877     & 0.962     & 0.828     & 0.567     \\
V4.5 & 0.825     & 0.896     & 0.993     & 0.884     & 0.815     & 0.828     & 0.890     & 0.971     & 0.827     & 0.537     & 0.821     & 0.865     & 0.971     & 0.827     & 0.537     \\
V4.6 & 0.823     & 0.884     & 0.992     & 0.875     & 0.823     & 0.795     & 0.862     & 0.968     & 0.812     & 0.557     & 0.868     & 0.870     & 0.968     & 0.812     & 0.557  \\ \hline  
\end{tabular}
\vspace{-3.0ex}
\end{table*}

\begin{table*}[htb]
\scriptsize
\caption{Optimum value of \textit{m}}
\label{mtable}
\centering
\begin{tabular}{ccc|cc|cc|cc}
\hline
\multicolumn{9}{c}{\textbf{DATASET-2}}                                                                                                                                                          \\ \hline
                     & \multicolumn{2}{c|}{\textbf{m/4}}        & \multicolumn{2}{c|}{\textbf{m/2}}        & \multicolumn{2}{c|}{\textbf{3m/4}}       & \multicolumn{2}{c}{\textbf{m}}           \\ \hline
                     & \textbf{TextConvoNet\_6} & \textbf{TextConvoNet\_4} & \textbf{TextConvoNet\_6} & \textbf{TextConvoNet\_4} & \textbf{TextConvoNet\_6} & \textbf{TextConvoNet\_4} & \textbf{TextConvoNet\_6} & \textbf{TextConvoNet\_4} \\ \hline
\textbf{Accuracy}    & 0.890           & 0.886           & 0.901           & 0.887           & 0.889           & 0.888           & 0.901           & 0.894           \\
\textbf{Precision}   & 0.894           & 0.864           & 0.903           & 0.901           & 0.888           & 0.875           & 0.899           & 0.911           \\
\textbf{Recall}      & 0.844           & 0.872           & 0.863           & 0.828           & 0.849           & 0.863           & 0.867           & 0.835           \\
\textbf{F1\_score}   & 0.868           & 0.868           & 0.882           & 0.863           & 0.868           & 0.869           & 0.883           & 0.871           \\
\textbf{Specificity} & 0.925           & 0.896           & 0.930           & 0.932           & 0.919           & 0.907           & 0.926           & 0.939           \\
\textbf{Gmean1}      & 0.869           & 0.868           & 0.883           & 0.864           & 0.868           & 0.869           & 0.883           & 0.872           \\
\textbf{Gmean2}      & 0.883           & 0.884           & 0.896           & 0.878           & 0.883           & 0.885           & 0.896           & 0.885           \\
\textbf{MCC}         & 0.775           & 0.768           & 0.798           & 0.769           & 0.773           & 0.771           & 0.798           & 0.784           \\
\hline
\multicolumn{9}{c}{\textbf{DATASET-4}}                                                                                                                                                \\ \hline
                     & \multicolumn{2}{c|}{\textbf{m/4}}        & \multicolumn{2}{c|}{\textbf{m/2}}        & \multicolumn{2}{c|}{\textbf{3m/4}}       & \multicolumn{2}{c}{\textbf{m}}           \\ \hline
                     & \textbf{TextConvoNet\_6} & \textbf{TextConvoNet\_4} & \textbf{TextConvoNet\_6} & \textbf{TextConvoNet\_4} & \textbf{TextConvoNet\_6} & \textbf{TextConvoNet\_4} & \textbf{TextConvoNet\_6} & \textbf{TextConvoNet\_4} \\ \hline
\textbf{Accuracy}    & 0.826           & 0.820           & 0.827           & 0.828           & 0.829           & 0.829           & 0.829           & 0.824           \\
\textbf{Precision}   & 0.565           & 0.551           & 0.569           & 0.570           & 0.573           & 0.572           & 0.572           & 0.559           \\
\textbf{Recall}      & 0.565           & 0.551           & 0.569           & 0.570           & 0.573           & 0.572           & 0.572           & 0.559           \\
\textbf{F1\_score}   & 0.565           & 0.551           & 0.569           & 0.570           & 0.573           & 0.572           & 0.572           & 0.559           \\
\textbf{Specificity} & 0.891           & 0.888           & 0.892           & 0.893           & 0.893           & 0.893           & 0.893           & 0.890           \\
\textbf{Gmean1}      & 0.565           & 0.551           & 0.569           & 0.570           & 0.573           & 0.572           & 0.572           & 0.559           \\
\textbf{Gmean2}      & 0.710           & 0.699           & 0.712           & 0.713           & 0.715           & 0.715           & 0.715           & 0.705           \\
\textbf{MCC}         & 0.457           & 0.438           & 0.461           & 0.463           & 0.466           & 0.465           & 0.465           & 0.449.           \\          
\hline
\end{tabular}
\vspace{-3.0ex}
\end{table*}

\begin{figure}[htb]
\vspace{-2.0ex}
\centering
  \includegraphics[width=2.5in,height=1.5in,angle=0]{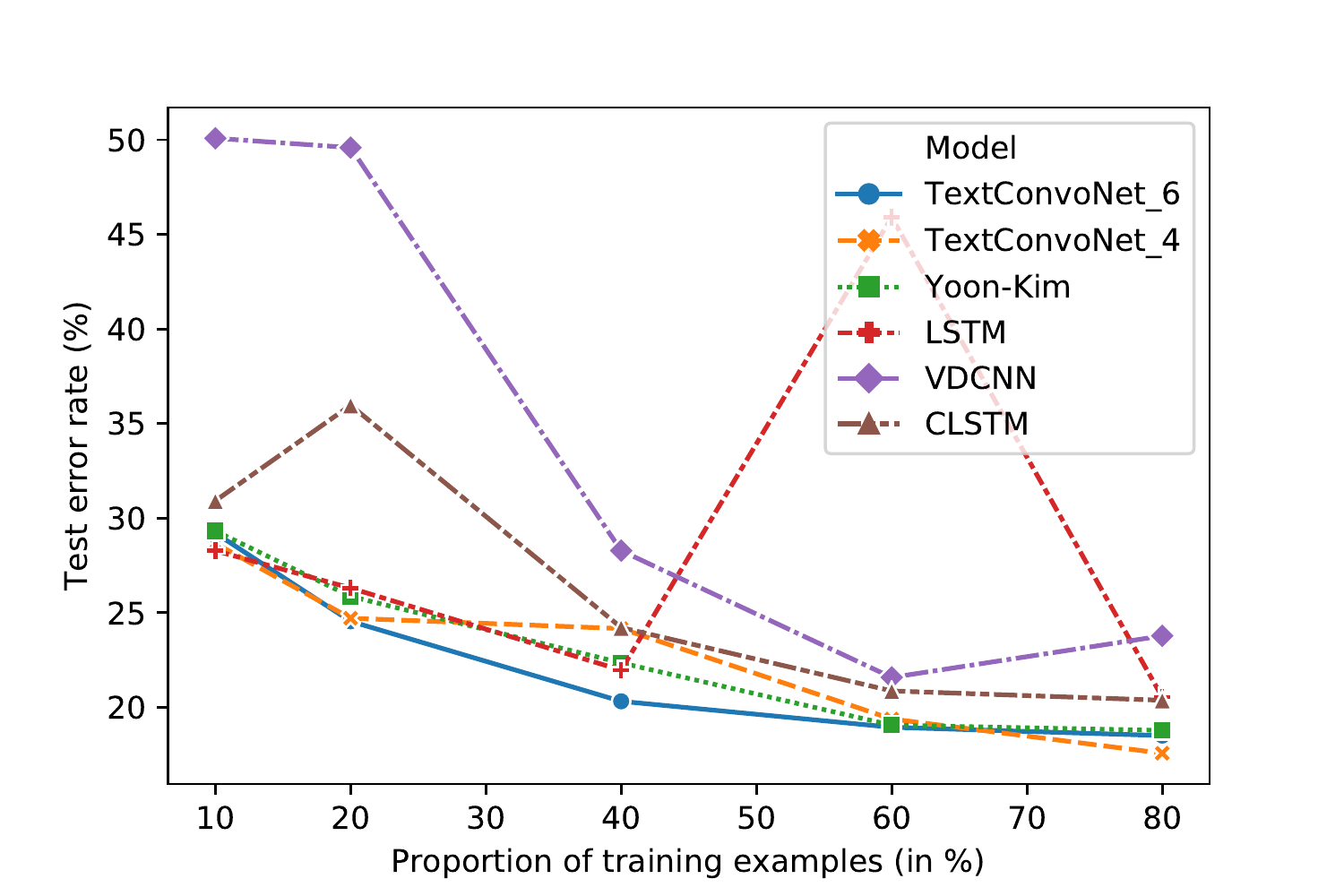}\quad
   \caption{Test error rates on Dataset-1 (Binary)}
    \includegraphics[width=2.5in,height=1.5in,angle=0]{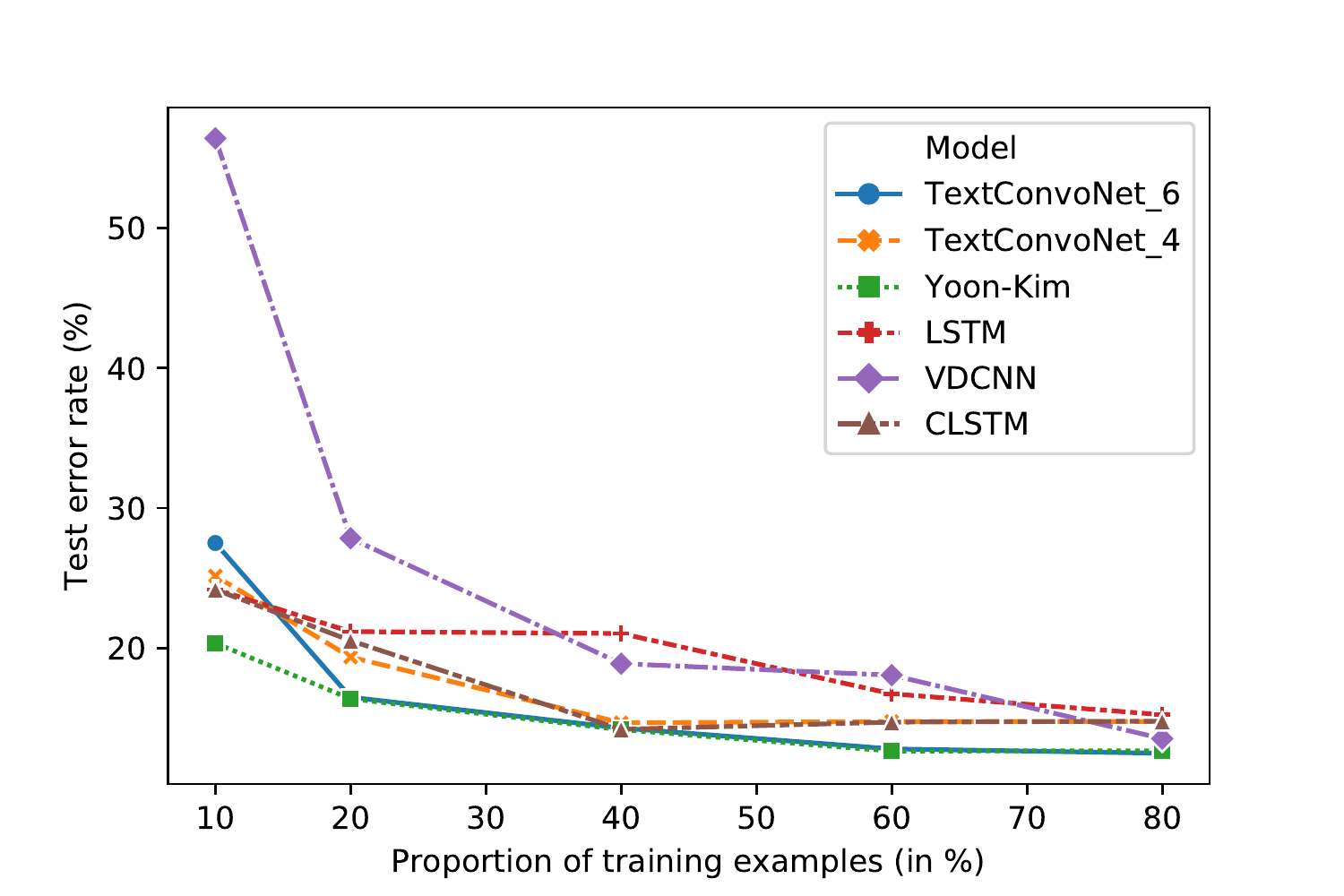}\quad
  \caption{Test error rates on Dataset-4 (Multi-class)}
  \vspace{-2.0ex}
\end{figure}

\subsection{Fewshot Learning}
In minimal data and challenging scenarios, it becomes rather important that the model can train well on a minimalist dataset (i.e., a dataset with a fewer number of training instances) and perform reasonably well on the test set \cite{yan2018few}. Fewshot learning for text classification is a scenario in which a small amount of labeled data for each category is available. The goal is for the prediction model to generalize to new unseen examples in the same categories both quickly and effectively. In the experiment, the test dataset's size remains constant as mentioned in Table~\ref{datasetdetails} for all the training percentages and is plotted against the test error rate at those training percentages. We evaluate \textit{TextConvoNet\_6}, \textit{TextConvoNet\_4}, Kim’s CNN model, LSTM, VDCNN, on one of the binary Dataset and one multi-class dataset with varying  proportions of training examples. The results are shown in Figures 4 and 5.

It is observed that the \textit{TextConvoNet} model performs better than all the other baseline models with lower test error rates at even lower proportion of training examples and \textit{TextConvoNet} was able to achieve it without any change in it's parameter space. \textit{TextConvoNet} extracts not just the n-gram based characteristics between the words of the same sentence as 1-D CNN does, but also the inter-sentence n-gram based features. As a result, \textit{TextConvonet} will be able to extract additional features that 1-D CNN models will not be able to.
This strengthens our claim that the proposed \textit{TextConvoNet} performs reasonably well even with a lesser number of training examples.

\section{Conclusion and future work}
In this paper, we presented a convolutional neural network-based deep learning architecture for text classification. The important feature of \textit{TextConvoNet} is that it not only extract the \textit{n-gram} features from the text data, but also capture the inter-sentence n-gram features. This was made possible by providing alternate input representation for text data. The extensive performance evaluation has shown that the proposed \textit{TextConvoNet} is effective for binary and multi-class classification problems. Therefore, it is evident that extracting the inter-sentence relationship improves the text classification task. 

In future work, we would explore the idea of representing input in higher dimensions so that convolution operations can capture various features from the textual data.  

\appendices
\section{}
We have prepared a supplementary file,  which includes the following details.  The first part discusses the implementation of all the machine learning and deep learning models with the used values of the control parameters. The next part contains additional results which were not reported in the paper due to space constraints.  This file is available in the GitHub repository\footnote{https://github.com/sonisanskar/TextConvoNet} and is uploaded with the paper.

\balance

\ifCLASSOPTIONcaptionsoff
  \newpage
\fi

\bibliographystyle{IEEEtran}
\bibliography{main.bib}

\balance

\balance

\end{document}